\documentclass[a4paper]{article}
\usepackage{amsmath, amssymb}
\usepackage[unicode=true]{hyperref}
\usepackage[ruled, noline]{algorithm2e}
\usepackage[numbers,compress]{natbib}
\usepackage[caption=false,font=footnotesize]{subfig}
\usepackage{fullpage}

\newcommand{\bb}[1]{\boldsymbol{#1}}
\newcommand{\minimize}{\operatorname*{minimize}}

\usepackage{graphicx}

\title{Fast and robust multiplane single molecule localization microscopy using deep neural network}
\author{Toshimitsu Aritake, Hideitsu Hino, Shigeyuki Namiki\\Daisuke Asanuma, Kenzo Hirose, Noboru Murata}

\begin{document}
\maketitle
\begin{abstract}
    Single molecule localization microscopy is widely used in biological research for measuring the nanostructures of samples smaller than the diffraction limit.
    This study uses multifocal plane microscopy and addresses the 3D single molecule localization problem, where lateral and axial locations of molecules are estimated. 
    However, when we multifocal plane microscopy is used, the estimation accuracy of 3D localization is easily deteriorated by the small lateral drifts of camera positions.
    We formulate a 3D molecule localization problem along with the estimation of the lateral drifts as a compressed sensing problem,
    A deep neural network was applied to accurately and efficiently solve this problem.
    The proposed method is robust to the lateral drifts and achieves an accuracy of 20~nm laterally and 50~nm axially without an explicit drift correction.
\end{abstract}
\section{Introduction}
Fluorescence microscopy is widely used in biological research to analyze in vivo structures of samples.
However, due to the diffraction limit of light, the resolution of conventional fluorescence microscopy is limited to approximately 200~nm laterally and 500~nm axially.
To overcome the diffraction limit, a number of super-resolution microscopy methods including single molecule localization microscopy (SMLM) have been proposed~\cite{Schermelleh2019}.
The fundamental problem of the super-resolution microscopy is to estimate the true molecular distribution from an observed image.
In SMLM, only a few molecules can be activated at a time by using photoactivatable molecules.
The positions of a few molecules can be accurately estimated by a localization algorithm such as Gaussian fitting.
By integrating the localization results of many frames, high-resolution images can be obtained.

%
In many biological studies, three-dimensional imaging techniques are important to observe the 3D structures of samples and various 3D fluorescence microscopy techniques have been proposed~\cite{Liu2018}.
In the past decade, SMLM methods have been extended to achieve 3D super-resolution.
The most commonly used method for 3D SMLM is point spread function (PSF) engineering. Several types of filters, such as astigmatism~\cite{Huang810}, double-helix~\cite{Pavani2995}, and teterapod~\cite{PhysRevLett.113.133902}, have been proposed to achieve 3D localization.
In these methods, the 3D molecule locations are estimated from the difference of the shape of the PSFs. 
However, when the molecule density is high, estimation errors of the axial position by these methods are high.
Also, when using the engineered PSFs, we need to use additional instruments, such as a cylindrical lens or phase masks, are required for the optical system.
Therefore, the reconfiguration of the optical system for other applications is difficult.

In this work, multifocal plane microscopy (MUM)\cite{Ram2008} is used for 3D SMLM.
MUM is a simple extension of 2D fluorescence microscopy to 3D by using multiple cameras.
In this study, quad-plane microscopy is used as MUM.
The 3D locations of the molecules are estimated by the images obtained from four focal planes.
This study did not use engineered PSFs for the microscopy work; thus, the optical layout is simple and the reconfiguration of the optical system is relatively easy.
Moreover, even when the molecule density is high, estimation errors are not as high as the engineered PSF because the shape of the PSF is simpler.
The major problem when using quad-plane microscopy for 3D SMLM is lateral drifts of the camera positions which affect the localization quality.
When we use quad-plane microscopy, the positions of cameras may have sub-pixel lateral drifts.
The location of the molecules is estimated from the observation for each focal plane; hence,  
the drifts of the camera positions make the estimation less accurate.

For an accurate localization of the molecules, it is necessary to estimate the amount of lateral drifts.
In this study, the 3D molecule location is estimated along with the amount of lateral drifts. This estimation problem is formulated as a compressed sensing problem~\cite{1614066}.
However, as the size of the input images become larger, the problem becomes intractable due to its high computational cost.

Recently, deep neural networks (DNNs) have received growing attention and have been successfully applied in a wide variety of applications 
due to their high predictive performances.
In the last few years, convolutional neural networks (CNNs), which is a type of a DNN, have been applied to SMLM~\cite{Zelger:18,Ouyang2018,Nehme:18,Boyd267096}.
CNNs have achieved a remarkable speedup of 2D and 3D molecule localization.
Although training a neural network takes several hours to days, the trained network can estimate the molecule location accurately and efficiently.
In addition, DNNs are suitable for SMLM because an infinite number of training data can be generated by using an approximated PSF.
This is because the performance of a neural network is significantly affected by the amount of data.

We also employ a CNN to efficiently estimate the location of molecules. 
The architecture of the proposed network is based on the fast super-resolution convolutional neural network (FSRCNN)~\cite{fsrcnn}, which was used for the super-resolution of natural images.
The network estimates the sub-pixel location of the molecules in an input image by predicting the molecules' existence at each location.
In addition, this network is trained to be robust to lateral drifts of camera positions so that the network can localize molecules accurately without explicit drifts correction.

The rest of this paper is organized as follows.
The experimental setting of a quad-plane microscope used in this study is presented in section 2,
and the formulation of the single molecule localization problem is explained in section~3.
Then the problem of the molecule localization using quad-plane microscopy and the detail of the proposed method are presented in section~4.
The experimental results validate the algorithm as shown in section~5.
Finally, the concluding remarks and the discussion are provided in the last section.

\section{Experimental setting}
\begin{figure}[t]
    \centering
    \includegraphics[width=.7\textwidth]{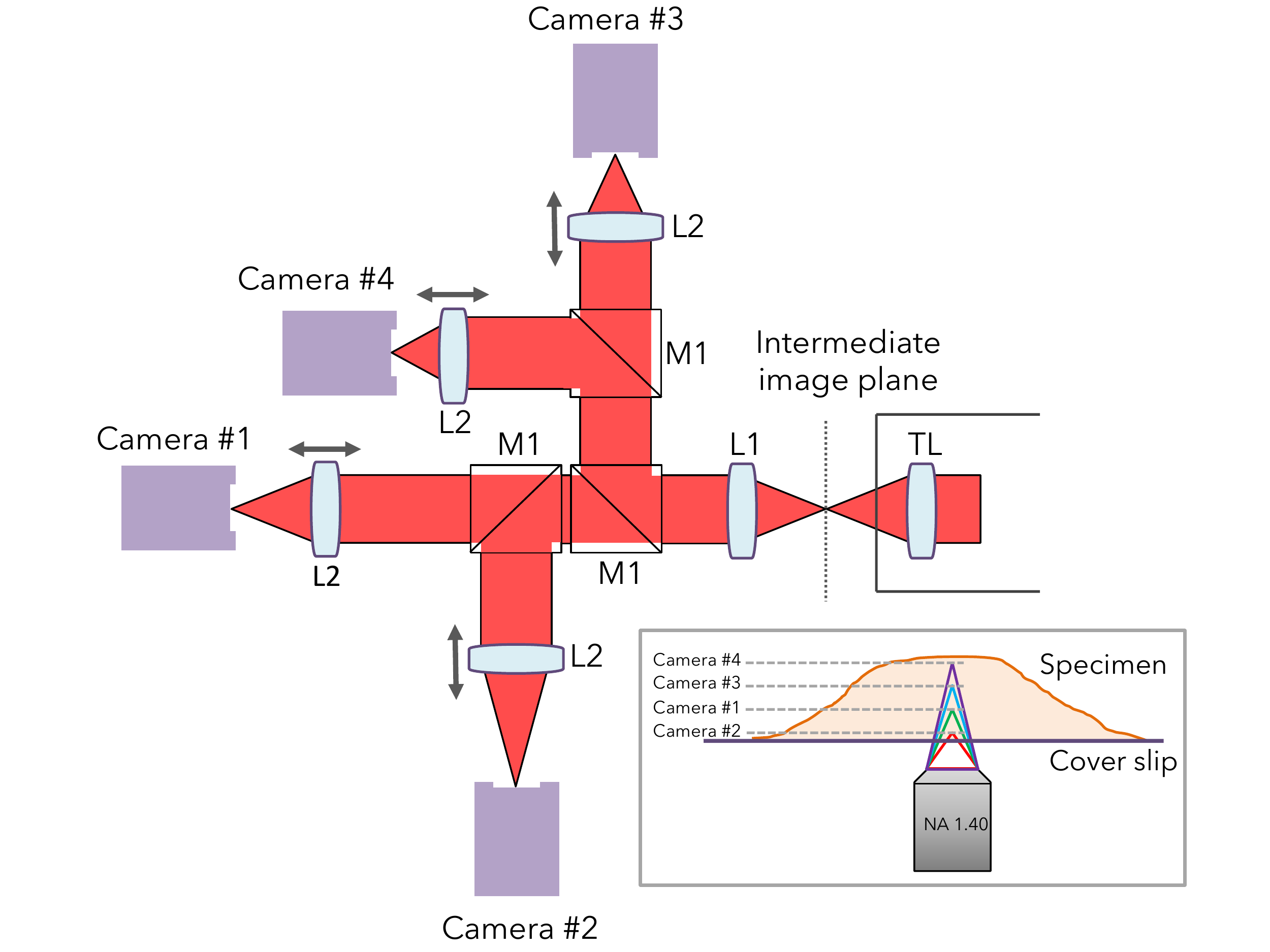}
    \caption{Optical layout of the quad-plane microscope. The intermediate image is relayed onto each camera via a pair of lenses (L1, f=125.0 mm; L2, f=100.0 mm). TL, tube lens; M1, 1:1 beam-splitter mirror. The inset shows the focusing planes of four cameras.}
    \label{fig:microscope}
\end{figure}

A multi-focus microscope equipped with four EM-CCD cameras (iXon 897, Andor) was constructed based on a commercial inverted microscope (ECLIPSE Ti, Nikon) (Fig. \ref{fig:microscope}).
A 640 nm laser beam (HL6366DG, Thorlabs) that passed through a cleanup filter (LD01-640/8, Semrock) was focused on the back focal plane of a 100× oil immersion objective (Plan Apo VC 100X/1.40, Nikon) to illuminate an Alexa Fluor 647-stained specimen at an excitation intensity of approximately 5 kW/cm${}^2$.
The fluorescence emitted from the specimen was collected by the same objective.
A filter cube consisting of an excitation filter (608--648 nm, FF02-628/40, Semrock), a dichroic mirror (669 nm, FF660-Di02, Semrock), and a bandpass mirror (672--712 nm, FF01-692/40, Semrock) was used to separate the excitation and emission light.
The fluorescence image formed by the internal tube lens of the inverted microscope was relayed by an achromatic lens (f = 125.0 mm, Thorlabs), split twice by 1:1 beam-splitter mirrors (BSW29R, Thorlabs), and refocused onto the four cameras via achromatic lenses (f = 100.0 mm, Thorlabs).
The axial positions of the achromatic lenses in front of the cameras were adjusted so that the four planes at 400 nm intervals in the Z-axis direction of the specimen correspond to the conjugate planes of the sensor surface of the respective camera.
The relative distance among planes was estimated by a shift in Z-axis position dependence of PSFs which are determined by imaging fluorescent beads (FluoSphere Carboxylate-Modified Microspheres, 0.2 $\mu$m, Invitrogen) while varying Z-axis positions of the objective by using a piezo positioner (P725.1, PI).
The difference in the field of view of the cameras was corrected by coordinate registration using affine transformation, parameters of which were determined by images of multiple fluorescent beads captured on the different cameras. 

Methanol-fixed COS7 cells were used for STORM imaging of tubulin molecules expressing inside the cells as described previously~\cite{doi:10.1146/annurev.bi.54.070185.001555}. The first and secondary antibodies were an anti-tubulin antibody (YL1/2, Abcam) and an Alexa Fluor 647-labeled anti-rat IgG antibody, respectively. The specimen was mounted in a STORM buffer (10 mM NaCl, 60\% sucrose, 10\% glucose, 0.1\% $\beta$-mercaptoethanol, 0.5 mg/mL glucose oxidase, 0.04 mg/mL catalase, and 50 mM HEPES, pH 8.0) and then subjected to imaging. Images were acquired at 22Hz with 20 ms exposure.
\section{Formulation}
In this work, we assume that the resolution of observed images and the target resolution is given.
Therefore, a target 3D space was divided into voxels. This was used to model the observation on a grid of voxels. 
Also, the molecules are localized at the voxel level of the specified resolution.

\subsection{Observation Model}

Let $\bb y^l=(y_1^l, y_2^l, \ldots ,y_n^l)\in\mathbb{R}^n$ be an observed low-resolution image obtained by quad-plane microscopy, where $y_i^l$ is the observed fluorescence intensity at the $i$-th observation coordinate $\bb x_i^l$.
Since $\bb y^l$ is a convolution of the true molecule density and the PSF, $\bb y^l$ can be approximated by a linear equation as,
\begin{equation}
    \bb y^l\approx \bb H \bb w,
\end{equation}
where $\bb H\in\mathbb{R}^{n\times m}$ is an observation matrix and $\bb w=(w_1, w_2, \ldots ,w_m)\in\mathbb{R}^m$ is a molecule distribution where $w_j$ represents the weight of the intensity of the molecule at the $j$-th voxel coordinates $\bb x_j^h$.
Here, $\bb x^l_i\in\mathbb{R}^3\ (i=1,2, \ldots, n)$ and $\bb x^h_j\in\mathbb{R}^3\ (j=1,2,\ldots ,m)$ are coordinates of the grid of the low- and high- resolution voxels respectively.
The $(i,j)$-element of the matrix $\bb H$ represents the fluorescence from a molecule at $\bb x^h_j$ observed at $\bb x^l_i$, and
can be written as $h(\bb x_i^l, \bb x_j^h)$, where $h$ is a point spread function:
\begin{equation*}
    \bb H =
    \left[
        \begin{array}{cccc}
            h(\bb x^l_1,\bb x^h_1) & 
            \cdots & h(\bb x^l_1,\bb x^h_m)\\
            h(\bb x^l_2,\bb x^h_1) & 
            \cdots & h(\bb x^l_2,\bb x^h_m)\\
            \vdots  & \ddots & \vdots\\
            h(\bb x^l_n,\bb x^h_1) & 
            \cdots & h(\bb x^l_n,\bb x^h_m)\\
        \end{array}
    \right]\in\mathbb{R}^{n\times m}.
\end{equation*}

We assume that an observed image contains both shot noise and additive observation noise.
The shot noise follows a Poisson distribution and the observation noise follows a Gaussian distribution for each observation independently.
Hence the observation can be modeled as
\begin{equation}
    \bb y^{l} = \bb H\bb w + \bb\epsilon.
    \label{eq:observation_model}
\end{equation}
where $\bb\epsilon$ is composed of shot noise and observation noise.

Now, the problem of molecule localization is to estimate weights $w_j$ for all $j=1,2,\ldots m$ from a low-resolution image $\bb y^l$.
Here, the observation matrix $\bb H$ is an overcomplete matrix $(n<m)$; thus the coefficient vector $\bb w$ cannot be recovered by minimizing the noise in Eq. \eqref{eq:observation_model}.
However, since $\bb w$ is a sparse vector, $\bb w$ can be recovered by solving the following linear inverse problem:
\begin{equation}
    \minimize_{\bb w} \|\bb y^l-\bb H \bb w\|_2^2 + \lambda \|\bb w\|_1,
    \label{eq:Lasso}
\end{equation}
which is known as Lasso (see \cite{Hastie:2015:SLS:2834535} and reference therein).
This type of inverse problem is known as compressed sensing~\cite{1614066}.
\subsection{Three-dimensional Point Spread Function}
In the above formulation, the true PSF $h$ is not known in general; therefore, we use a parametric function $\hat{h}$ to approximate $h$ and an approximated observation matrix $\hat{\bb H}$ is used to solve Eq.~\eqref{eq:Lasso}.
In this work, we use quad-plane microscopy which takes four images at different degrees of defocus.
The width and the peak of the $h(\bb x^l_i, \bb x^h_j)$ depend on the distances between a molecule and the focal planes.

\begin{figure}[t]
    \centering
    \includegraphics[width=.7\textwidth]{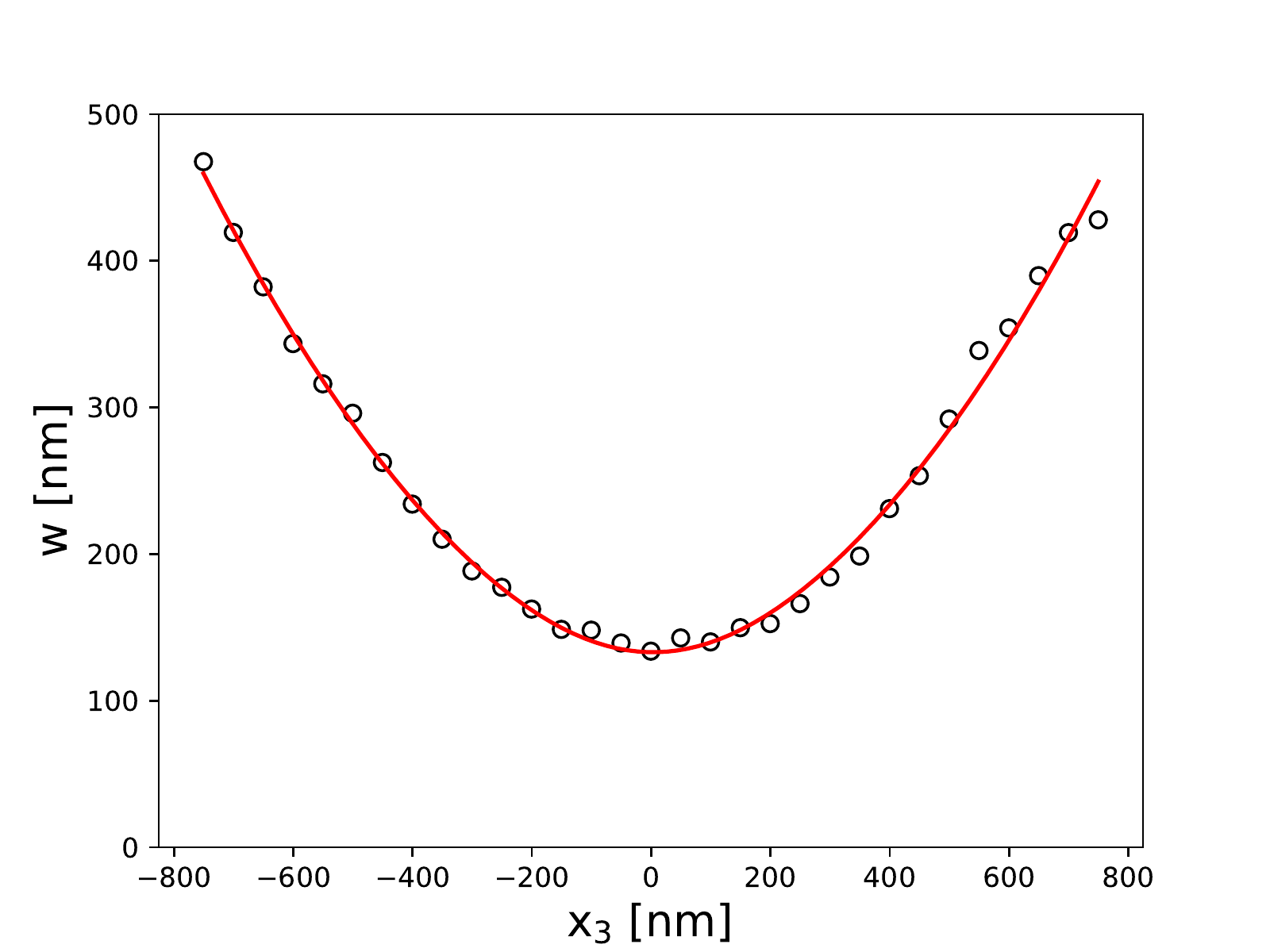}
    \caption{The width of the observed fluorescent beads and values of the defocus curve. The observed width of the fluorescent beads at each depth are shown by a circle. The red line shows the value of the defocus curve to approximate the width with the parameters in the Table. \ref{tab:parameters}.}
    \label{fig:defocus_curve}
\end{figure}

The PSF of the quad-plane microscopy is modeled by the following function:
\begin{align}
        &\hat{h}(\bb x^l_i, \bb x^h_j)\nonumber\\
        &=\hat{h}(x_{i1}^l, x_{i2}^l, x_{i3}^l, x_{j1}^h, x_{j2}^h, x_{j3}^h)\nonumber\\
        &= a(x_3)\exp\left(-\frac{(x_{i1}^l-x_{j1}^h)^2+(x_{i2}^l-x_{j2}^h)^2}{2(w(x_{i3}^l-x_{j3}^h))^2}\right)+b,
        \label{eq:PSF}
\end{align}
where $b$ is the background fluorescence.
This PSF is similar to the PSF used in \cite{Gu2014} for biplane microscopy.
The width $w(x_3)$ of the PSF varies depending on the axial position $x_3$ and is described by the following defocus curve:
\begin{equation}
    w(x_3)=w_0\sqrt{1+\left(\frac{x_3}{d}\right)^2+A\left(\frac{x_3}{d}\right)^3+B\left(\frac{x_3}{d}\right)^4},
    \label{eq:defocus_curve}
\end{equation}
where $w_0$ is the width of the PSF when a molecule is on the focal plane and $d$ is the focus depth of the microscope.
The peak $a$ of the PSF depends on the width $w(x_3)$ and is modeled as:
\begin{equation}
    a(x_3) = \frac{a'}{2\pi w(x_3)^2}.
\end{equation}
The parameters $a'$, $b$, $d$, $A$, $B$ are determined as listed by Table \ref{tab:parameters} using a set of images of fluorescent beads obtained from different depths. 
Figure \ref{fig:defocus_curve} shows the width of the observed fluorescent beads and the value of the defocus curve \eqref{eq:defocus_curve}.

\begin{table}[b]
    \centering
    \caption{Parameters of the PSF}
    \label{tab:parameters}
    \begin{tabular}{cc}
        \hline
        Parameters & Value\\
        \hline
        $a'$ & $5.00\times 10^7$\\
        $b$ & 0\\
        $w_0$ & $1.33\times 10^2$\\ 
        $d$ & $3.02\times 10^2$\\ 
        $A$ & $7.37\times 10^{-4}$\\
        $B$ & $6.27\times 10^{-3}$\\
        \hline
    \end{tabular}
\end{table}

\subsection{Lateral Drifts of the Focal Planes}
When we use multi-focal plane microscopy to localize the molecules, we need to consider the lateral drifts of focal planes.
Otherwise, the estimation accuracy of the localization is easily deteriorated, because the elements of the true $\bb H$ varies depends on the drifts.

Let $\Delta_{z1}$ and $\Delta_{z2}\in\mathbb{N}\ (z=1,2,3,4)$ be the amount of lateral drifts along the horizontal and vertical axis of the $z$-th focal plane.
Then, the drift vector is written as $\bb \Delta_z \equiv (\Delta_{z1}, \Delta_{z2}, 0)$.
In this work, we only consider the high-resolution voxel-level lateral drifts. 
The PSF was defined as Eq.~\eqref{eq:PSF} and the following equation holds for $\hat{h}(\bb x_i^l, \bb x^h_j)$:
\begin{equation}
    \hat{h}(\bb x_i^l+\bb\Delta_z,\bb x^h_j)=\hat{h}(\bb x_i^l, \bb x^h_j-\bb\Delta_z),
    \label{eq:lateral_shift}
\end{equation}
where the observation coordinates $\bb x^l_i$ is on the $z$-th plane.

This equation implies that the observation from the molecule at $\bb x^h_j$ with a lateral drift $\bb\Delta_z$ is identical to the 
observation from the molecule at $\bb x^h_j-\bb\Delta_z$ without a lateral drift.
Since the observation from each plane is affected by a different drift of camera position, the location of a molecule can be estimated as different points from each plane.
However, the problem Eq.~\eqref{eq:Lasso} considers all of the focal planes at the time to estimate the molecule location.
Hence, unless we know the amount of lateral drift for each plane, 
the true molecule position cannot be correctly estimated even when a single molecule exists in the 3D space.

Besides, based on Eq.~\eqref{eq:lateral_shift}, if $\bb\Delta_z$ is the same for all $z=1, 2, 3, 4$, we cannot distinguish if the amount of lateral drifts are $\bb \Delta_z$ or if the true molecule location is $\bb x_j^h-\bb\Delta_z$ even when we use the information of all planes.
Instead, we estimate the relative lateral drifts $\bb\Delta'_z=\bb\Delta_z-\bb\Delta_1\ (z=2, 3, 4)$ from a reference plane $z=1$.
In this study, the amount of relative lateral drifts, as well as the molecule potisions, are estimated at a high-resolution voxel level.

\section{Method}
\subsection{Compressed Sensing with Lateral Drift Estimation}
When solving the molecule localization problem stated in Eq.~\eqref{eq:Lasso}, it is necessary to consider the lateral drifts of the focal planes to estimate the molecule location accurately.
The approximated observation matrix $\hat{\bb H}$ varies depending on the lateral drifts; hence,the approximated observation matrix $\hat{\bb H}$ can be modified so that the molecule location is correctly estimated.

By ordering the rows of the observation matrix $\hat{\bb H}$ based on the axial position of the observation coordinate $\bb x_i^l\ (i=1,2,\ldots n)$,
$\hat{\bb H}$ becomes a block matrix
\begin{equation}
    \hat{\bb H} = \left[
        \begin{array}{c}
            \hat{\bb H}_1\\
            \hat{\bb H_2}\\
            \hat{\bb H_3}\\
            \hat{\bb H_4}
        \end{array}
    \right],
\end{equation}
where each submatrix $\hat{\bb H}_z\in\mathbb{R}^{\frac{4}{n}\times m}$ represents the observation matrix of the $z$-th focal plane.
Since the lateral drift affects each plane independently, the drift of each block can be considered individually.
Moreover, the absolute drift cannot be estimated from observation.
Therefore, we consider the plane $z=1$ as a reference plane. As a result, the relative drifts of $\hat{\bb H}_2$, $\hat{\bb H}_3$, and $\hat{\bb H}_4$ are considered.

Shifted submatrices with a shift $\bb \Delta_z'\ (z=2,3,4)$ are written as follows,
\begin{align}
        &\hat{\bb H}_z(\bb \Delta_z')\nonumber\\ \equiv&
        \left[
            \begin{array}{cccc}
                \hat{h}(\bb x^l_1-\bb\Delta_z',\bb x^h_1) & 
                \cdots & \hat{h}(\bb x^l_1-\bb\Delta_z',\bb x^h_m)\\
                \hat{h}(\bb x^l_2-\bb\Delta_z',\bb x^h_1) & 
                \cdots & \hat{h}(\bb x^l_2-\bb\Delta_z',\bb x^h_m)\\
                \vdots  & \ddots & \vdots\\
                \hat{h}(\bb x^l_{\frac{n}{4}}-\bb\Delta_z',\bb x^h_1) & 
                \cdots & \hat{h}(\bb x^l_{\frac{n}{4}}-\bb\Delta_z',\bb x^h_m)\\
            \end{array}
        \right],
        \label{eq:shifted_submatrix}
\end{align}
while considering the high-resolution pixel-level relative shift $\bb\Delta_z'$ and the maximum amount of shift as a hyperparameter.

Now, the problem is to estimate both the lateral drifts and the molecule locations from the observation $\bb y^{l}$, which can be formulated as:
\begin{align}
    \minimize_{\{\bb w_t\}_{t=1}^{T}, \{\bb \Delta_z'\}_{z=2}^4} \sum_{t=1}^T &\left(\|\bb y_t^{l}-\hat{\bb H}(\{\bb \Delta_z'\}_{z=2}^4)\bb w_t\|_F^2+\lambda \|\bb w_t\|_1\right),
    \label{eq:optimization_problem}
\end{align}
where 
\begin{equation}
    \bb H(\{\bb\Delta_z'\}_{z=2}^4) = \left[
        \begin{array}{c}
            \hat{\bb H}_1\\
            \hat{\bb H}_2(\bb\Delta_2')\\
            \hat{\bb H}_3(\bb\Delta_3')\\
            \hat{\bb H}_4(\bb\Delta_4')
        \end{array}
    \right].
\end{equation}
Here, we consider $T$ images at the same time so that the amount of the lateral drifts are correctly estimated from the images.
Since we assume that only a small number of molecules exist in the target 3D space, it is not always possible to estimate all of the lateral drifts from a single image.

However, the optimization problem above requires a high computational cost since we need to consider all of the possible pairs of lateral drifts to get an optimal solution.
Although this problem can be solved by alternating the optimization of $\{\bb w\}_{t=1}^{T}$ and $\{\bb \Delta_z'\}_{z=2}^4$ to get a sub-optimal solution,
the optimization of $\{\bb w_t\}_{t=1}^T$ still requires a high computational cost as the input image size becomes larger or the target resolution becomes higher.
Therefore, a faster method to solve the optimization problem is required to obtain a super-resolution image within a reasonable computational time.

\subsection{Convolutional Neural Network (CNN)}
In this work, we use a CNN to increase the computational speed to solve the problem Eq. \eqref{eq:optimization_problem}.
The molecule localization problem Eq. \eqref{eq:Lasso} is a deconvolution problem, where a molecule location is estimated from the observed images.
This deconvolution process can be seen as a composition of upsampling and deblurring of a low-resolution image.
In this work, the resolution of the input and output images is given and the scaling factor is $8\times$ along each axis.

\begin{figure}[t!]
    \centering
    \includegraphics[width=0.9\textwidth]{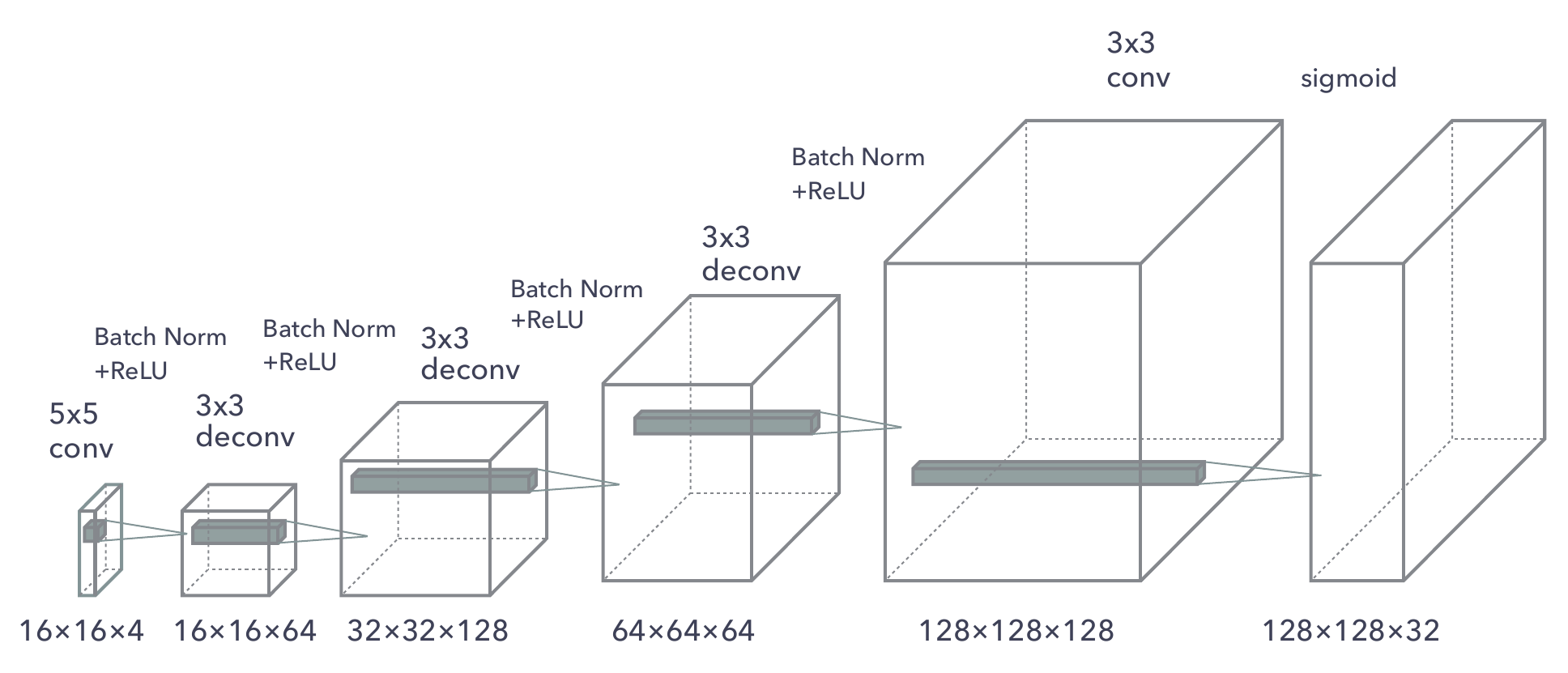}
    \caption{Network architecture of the proposed model.}
    \label{fig:architecture}
\end{figure}

This network is composed of several convolution layers and deconvolution layers as shown in Fig.~\ref{fig:architecture} and the structure is similar to the structure of the fast super-resolution convolutional neural network (FSRCNN)~\cite{fsrcnn}, which is used for a single image super-resolution of natural images.
The first layer extracts features from an input image and this is followed by three deconvolution layers.
The sampling frequency of the input image is doubled at each deconvolution layer while important features for the localization are extracted.

We use ReLU as activation functions followed by batch normalization layers~\cite{Ioffe:2015:BNA:3045118.3045167} to enhance the training speed and the estimation quality for these layers.

For the last layer of the network, a convolution layer is used to obtain a set of images for the target axial resolution.
Also, unlike a natural image super-resolution task, the purpose of the localization problem Eq. \eqref{eq:Lasso} is to estimate the existence of a molecule for each high-resolution voxel.
Hence, the network directly outputs the probability of the existence of a molecule in each voxel using a sigmoid function.
Namely, at the last layer of the network, the binary classification problem is solved for each voxel.

Let $\bb p=(p_1, p_2, \ldots p_m)$ be the ground-truth molecule existence probability for each voxel where:
\begin{equation}
    p_j = \begin{cases}
        0 & w_j \leq 0,\\
        1 & w_j > 0,
    \end{cases}
    \label{eq:target_value}
\end{equation}
and $\bb q = (q_1, q_2, \ldots ,q_m)$ be the molecule existence probability estimated by the CNN.
Since the network solves the binary classification problem at each voxel,
we use the sum of the binary cross-entropy (BCE):
\begin{equation*}
    \ell(\bb p, \bb q) = \sum_{j=1}^m \left(- p_j\log(q_j)-(1-p_j)\log(1-q_j)\right),
\end{equation*}
as the loss functions to train the network.

To train the network, we need a training dataset.
However, the true molecule density of the real samples is not known.
Therefore, we use artificial observed images generated from artificial distributions to train the network.

We generate random molecule distributions that contain $K$ molecules in a 3D space.
In this work, $K=3$ was used and the size of the target 3D space is 3072~nm~$\times$~3072~nm~$\times$~1200~nm and the coordinates of the molecules $\bb x_k^h\ (k=1,2,\ldots ,K)$ are independently drawn from a uniform distribution on this space.
The weight $w_k\ (k=1,2,\ldots , K)$ of each molecule is also independently drawn from a continuous uniform distribution on $[0.3, 1.0]$, and the
target value $\bb p$ is generated as Eq. \eqref{eq:target_value}.

In this work, we assume that the size of the low-resolution voxels is 192~nm~$\times$~192~nm~$\times$~400~nm and the size of a low-resolution image is $16\times 16\times 4$. The size the high-resolution voxel is 24~nm~$\times$~24~nm~$\times$~50~nm and the size of a high-resolution image is $128\times 128\times 32$.
The relative lateral drift of the focal planes $\Delta_{z1}'$, $\Delta_{z2}\ (z=2, 3, 4)$ are randomly chosen as $24d$~nm independently where $d$ is drawn from a discrete uniform distribution on $[-2, 2]\subset\mathbb{N}$.
Then, low-resolution images are generated by calculating the values of $\sum_{k=1}^K \hat{h}(\bb x, \bb x_k^h)$ on the low-resolution grids $\bb x = \bb x^l_i\ (i=1, 2,\ldots ,n)$.

To generate a training dataset, the coordinates of the molecules $\{\bb x_k^h\}_{k=1}^K$ and the relative lateral drifts of the focal planes $\{\Delta_{z1}\}_{z=2}^4$, $\{\Delta_{z2}\}_{z=2}^4$ are randomly drawn as above for each frame $t=1, 2, \ldots ,T$ independently.
By training the network with this dataset, the trained network is expected to become robust to the lateral drifts of camera positions within $[-2, 2]$ high-resolution voxels.

\section{Experiments}
We show experimental results of the localization by the proposed method with both artificial images and the real microscopy images.
Below, the experiments were performed on a NVidia Tesla V100 32GB GPU.

\subsection{Experiments with Artificial Images}

We used 90,000 low-resolution images $\{\bb y_t^l\}_{t=1}^{90000}$ and corresponding molecule existence probability $\{\bb p_t\}_{t=1}^{90000}$ generated from the random molecule distribution to train the neural network.
In addition, we use 10,000 test data were generated in the same way as the training dataset.
We use Adam~\cite{DBLP:journals/corr/KingmaB14} as an optimizer, where its parameters are $\beta_1=0.9$, $\beta_2=0.99$ and the initial learning rate is set to $1.0\times 10^{-3}$,
and the batch size is 100.
The epoch number of the optimization is 30, and dataset is shuffled at the end of each epoch.

\begin{figure}[b!]
    \centering
    \includegraphics[width=.7\textwidth]{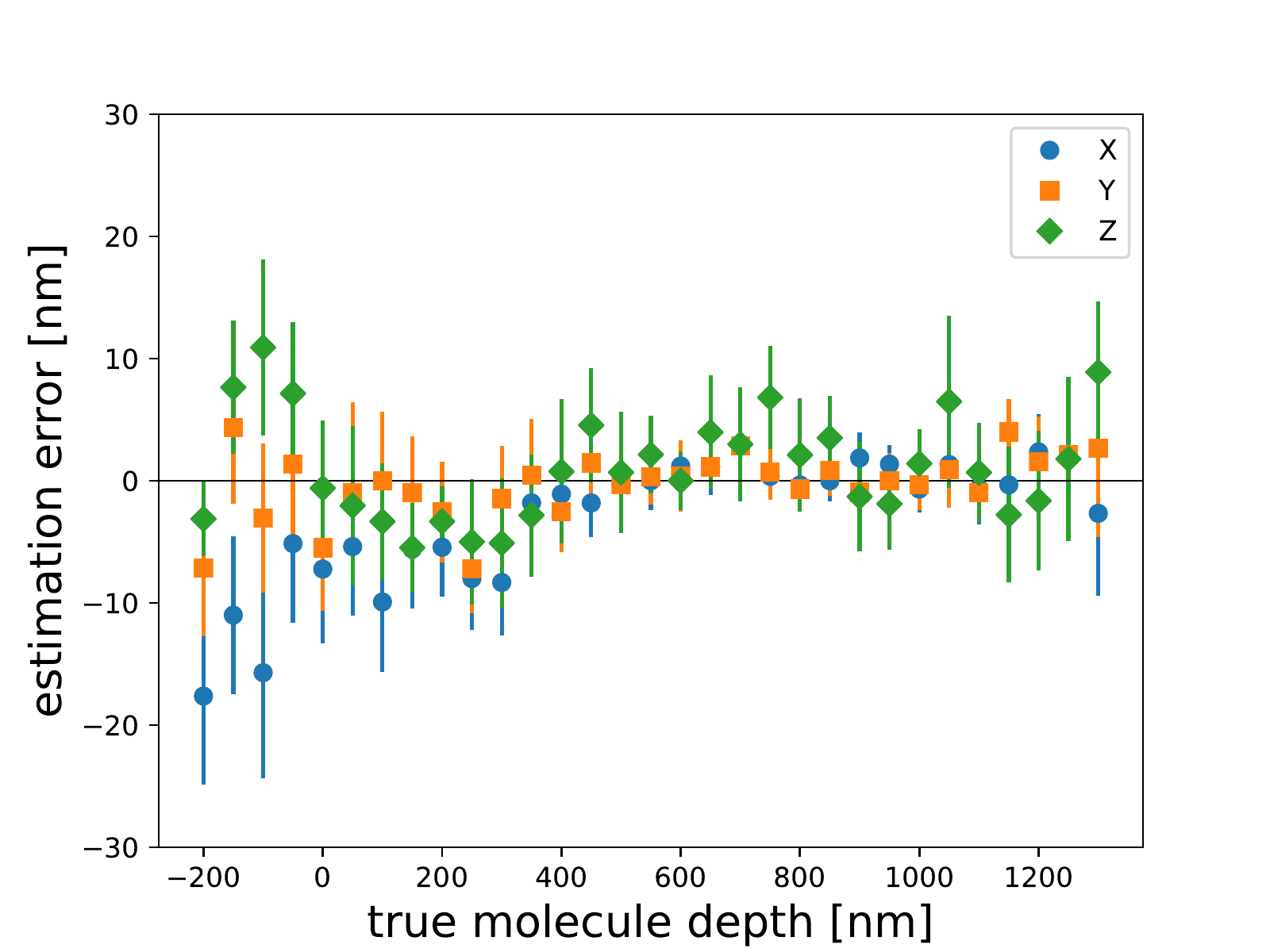}
    \caption{Average estimation error along the horizontal (X), vertical (Y), and axial (Z) axis and their 95\% confidence interval.}
    \label{fig:estimation_error}
\end{figure}

To validate the accuracy of the estimation by the trained network, we generated artificial images that contain only one molecule in the 3D space.
In this experiment, we estimate that a molecule exists in a voxel where the network outputs the highest molecule existence probability.
In Fig.~\ref{fig:estimation_error}, the mean localization accuracy along the horizontal (X), vertical (Y) and axial (Z) directions with 95\% confidence intervals are shown for each true molecule depth.
The figure indicates that the error along each axis is within a high-resolution voxel on average along each axis at all depth.

\begin{figure}[tb]
    \centering
    \subfloat[][Estimation result of $t=6$]{\includegraphics[width=.31\textwidth]{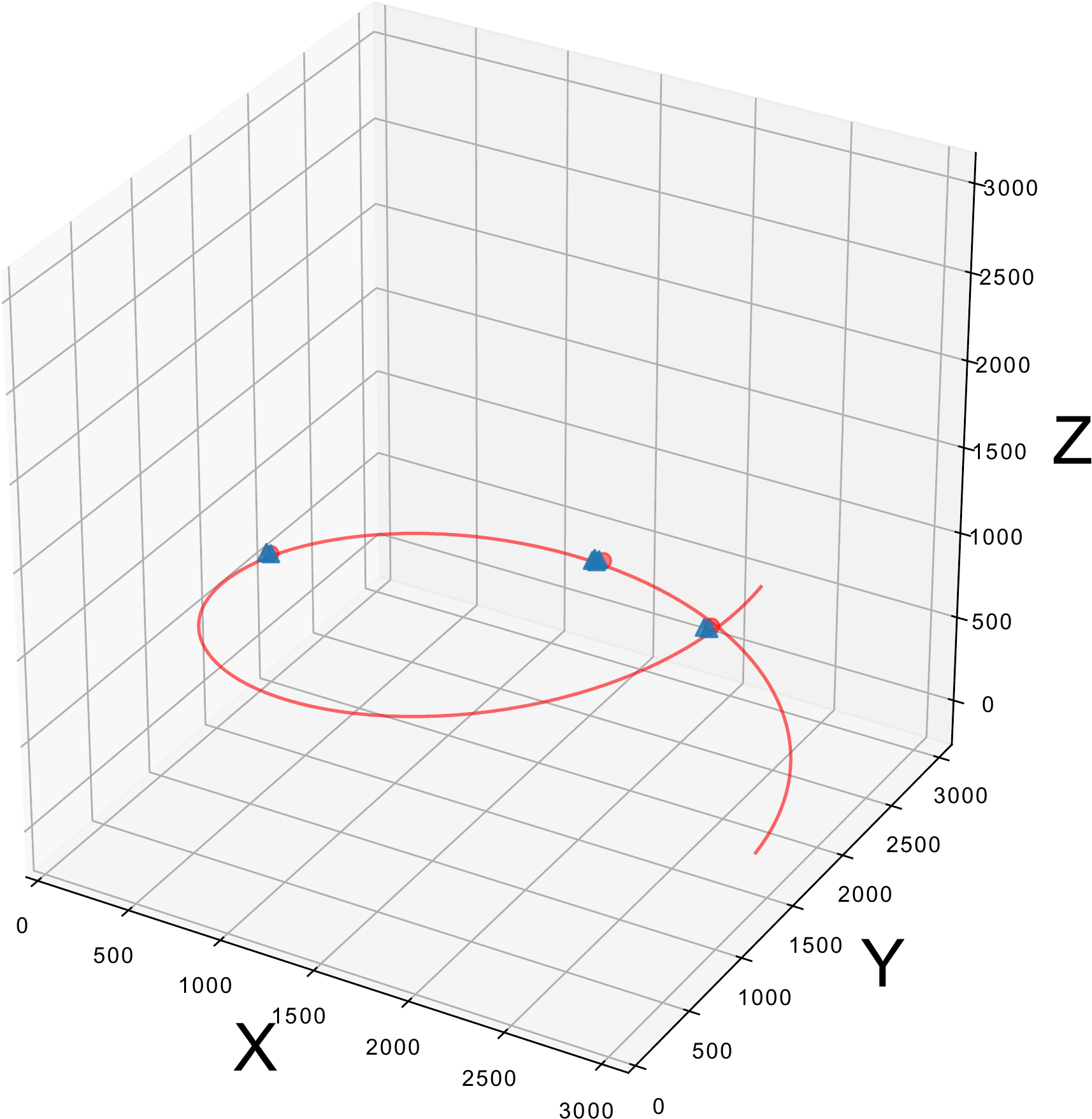}\label{fig:helix_a}}\quad
    \subfloat[][Estimation result of $t=43$]{\includegraphics[width=.31\textwidth]{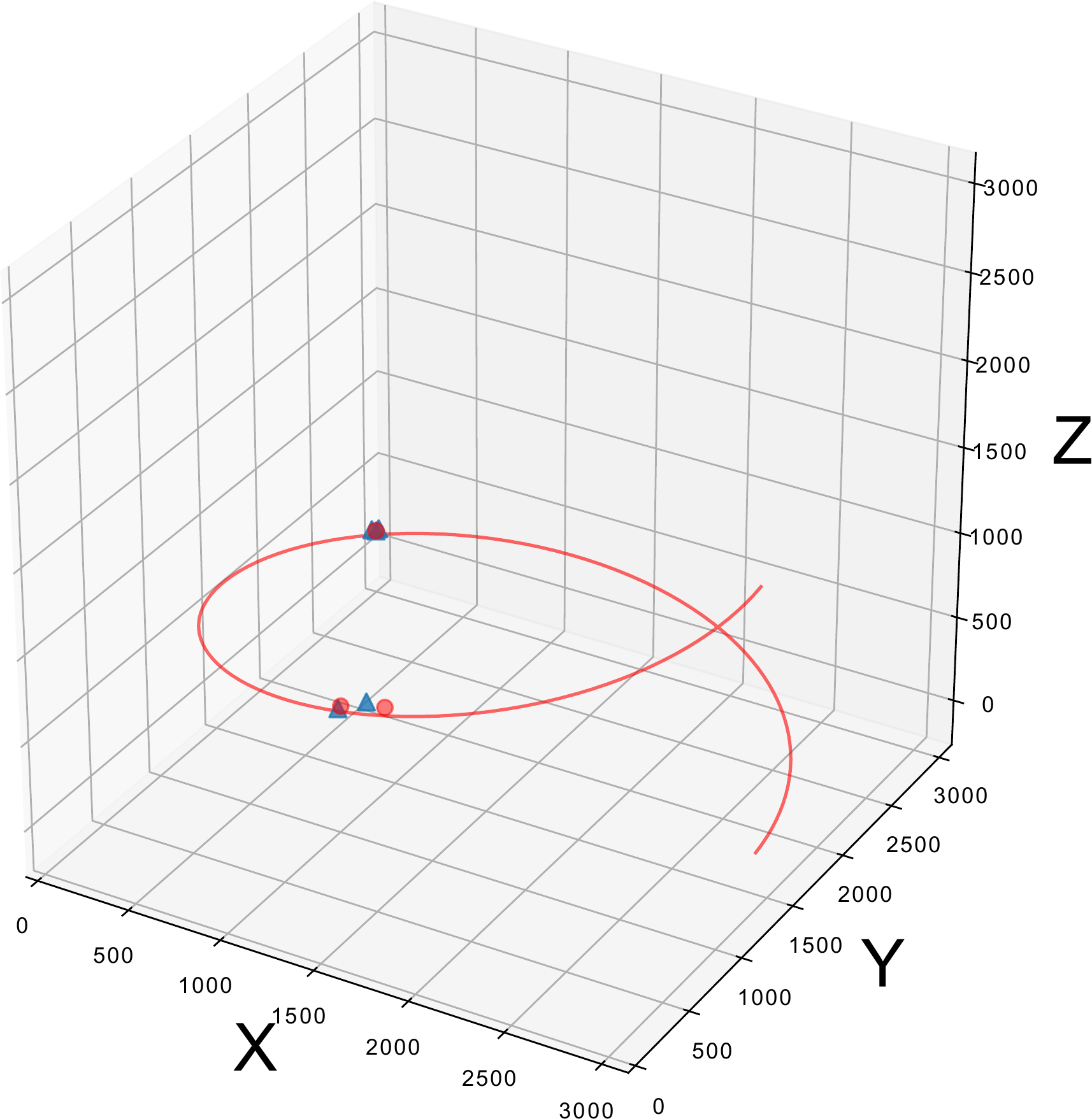}\label{fig:helix_b}}\quad
    \subfloat[][Reconstructed image from 300 frames]{\includegraphics[width=.31\textwidth]{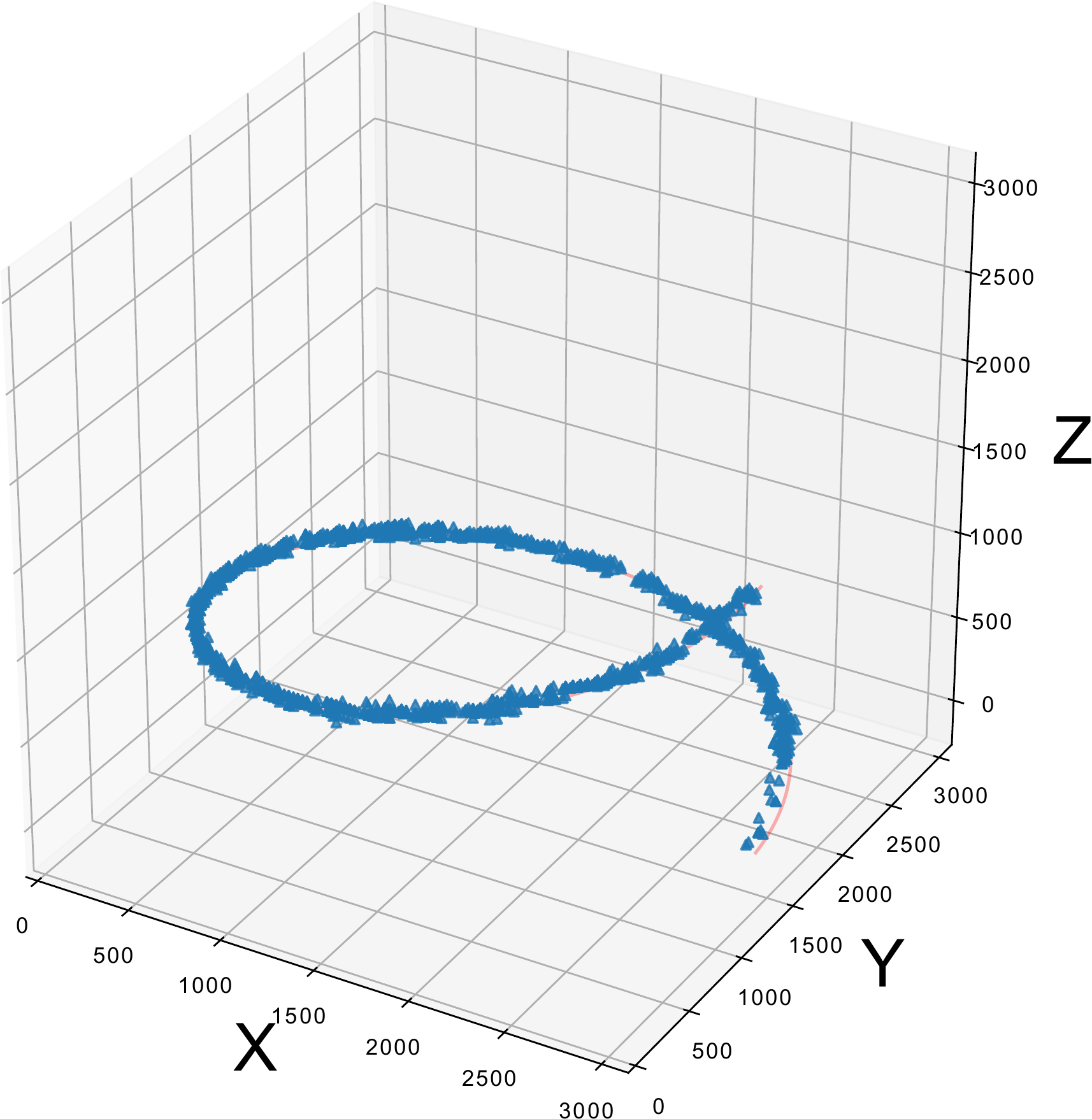}\label{fig:helix_c}}
    \caption{The localization result of artificial data. The red line shows the true molecule structure where molecules are sampled from. The red circles show a true high-resolution molecule coordinate and the blue triangles show a estimated molecule locations.
    The figure (a), (b) shows the estimation results of selected frames and the figure (c) shows the reconstructed image from 300 frames.}
    \label{fig:helix}
\end{figure}

Figure~\ref{fig:helix} shows the estimation results with multiple molecules.
The molecules are sampled from the helix curve (red line) and their high-resolution coordinates are shown by red circles.
In this experiment, we estimate that the molecules exist in the voxels whose molecule existence probability exceeds a certain threshold.
The thresholding value is common for all locations and needs to be specified by a user.
Here, we chose 0.1 for the thresholding value.
As demonstrated in Fig.~\ref{fig:helix}(a), three molecules distributed in the 3d space are also accurately detected.
Although, as Fig.~\ref{fig:helix}(b) indicates, the closely located molecules are difficult to localize, still, the estimated locations are close to the ground-truth locations
By plotting all of the detected molecules from all frames, the helix curve structure behind the molecules can be seen as the Fig.~\ref{fig:helix}(c).

\begin{figure}[t!]
    \centering
    \includegraphics[width=.7\textwidth]{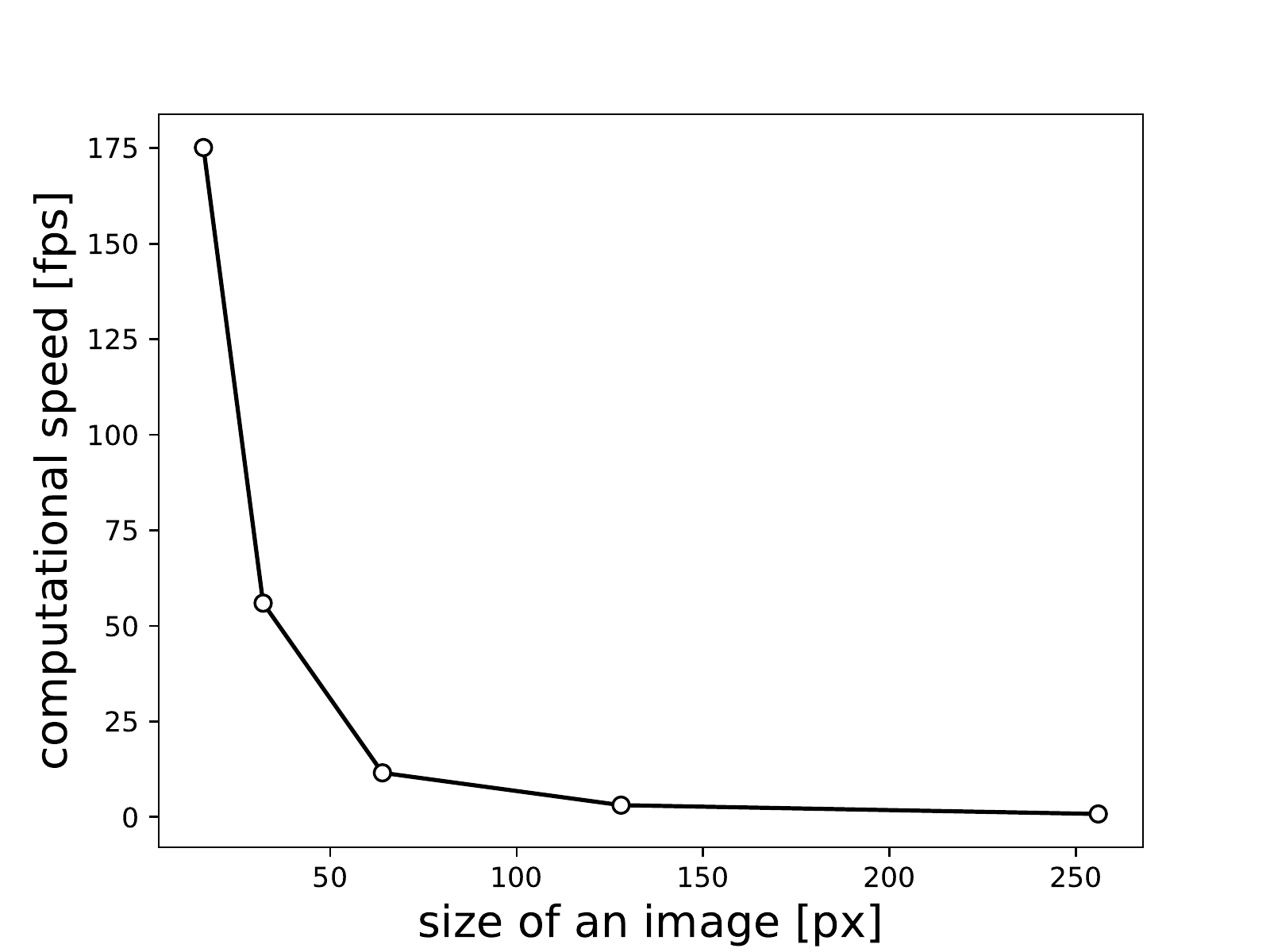}
    \caption{Computational speed (fps) of estimations by the trained network for images of the size 16 px $\times$ 16 px, 32 px $\times$ 32px, $\ldots$ ,256 px $\times$ 256 px. The estimation speed of our network is inverse proportional to the the number of pixels of an image.}
    \label{fig:computational_speed}
\end{figure}

The processing speed of the network is presented in Fig.~\ref{fig:computational_speed}.
As the figure indicates, the computational speed decreases as the size of the image increases and is inversely proportional to the number of pixels of an image.
Since we assume that 50 images are obtained by our microscopy at every second, further improvement of the processing speed is needed to process the large images in real-time.
Still, the computational time is significantly reduced in comparison to the compressed sensing method.
By solving the problem Eq. \eqref{eq:optimization_problem} by an alternating minimization, the processing speed is only $1.45\times 10^{-3}$ fps even for a small $16\times 16\times 4$ input.

\subsection{Experiments with Real Images}
In this section, we show the experimental result with real data that observed a microtubules by our microscopy.
In this experiment, there were no ground-truth results; hence, we used the trained neural network in the previous subsection to localized molecules.

The resolution of the low- and high-resolution image is the same as the previous subsection.
The size of input images for each frame is $256\times 256\times 4$ and the target image size is $2048\times 2048\times 32$.
The dataset contains 39,000 frames of images and each frame is processed independently to localize the molecules.
In this experiment, we estimate that a molecule exists in a voxel if the molecule's existence probability exceeds $0.05$.

\begin{figure}[b!]
    \centering
    \includegraphics[width=.7\textwidth]{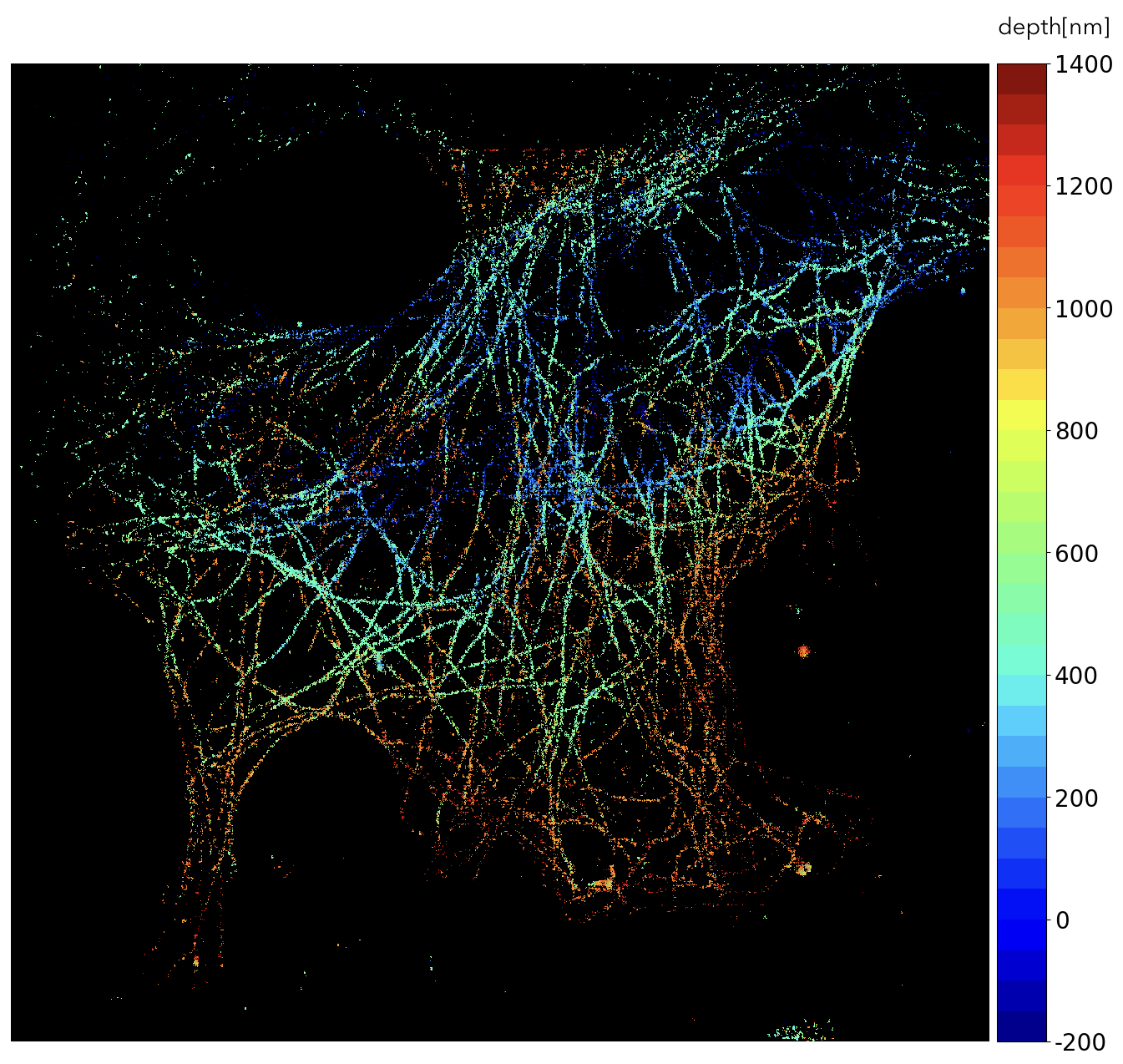}
    \caption{Estimated high-resolution image of the microtubules data. The depth-dependent tubular structure of the sample is visualized by the colors.}
    \label{fig:highres_tubulin}
\end{figure}

Figure~\ref{fig:highres_tubulin} shows the estimated high-resolution image at the selected depths generated by merging localization results of all of the frames.
Each pixel of the image is a binary value, which indicates that the voxel contains a molecule in more than one frame.
From the figures, we can see a tubular structure of the microtubules that varies depending on the depth.

\section{Discussion}
This study presents the 3D molecule localization problem using quad-plane microscopy.
The problem with using multi-focal plane microscopy (MUM) is that lateral drifts of camera positions make the localization less accurate.
We formulated the localization problem as a compressed sensing problem that consists of the molecule localization and an estimation of the amount of drifts.
However, the computational cost to solve this problem is high and the optimal solution cannot be obtained within a reasonable computational time.
A CNN is proposed to solve this problem accurately and efficiently. 
The network is trained to be robust against the sub-pixel lateral drifts for the camera locations.

The experiments with both artificial data and real data were presented.
The results suggest that the network achieves 3D localization of the molecules with a lateral resolution of 25~nm and an axial resolution of 50~nm on average.
It is also robust to the lateral drifts of the camera positions.
We expect this technique can be used to broaden the applicability of MUM for 3D imaging since an explicit drift correction is not required.

However, some limitations are worth noting.
Although, our proposed method significantly increased the computational speed of solving the localization problem,
it is still difficult to process large images with real-time processing speed.
Future work should, therefore, include further improvement in computational speed.
Using a faster method to extract possible molecule locations and localizing the molecule by the proposed method may further improve the computational efficiency.

\textbf{Acknowledgement}
The authors would like to thank M. Tanaka for technical assistance.
This work was partly supported by JSPS KAKENHI Grant Numbers 17H01793, 18H03291 and JST CREST Grant Number JPMJCR1761, JPMJCR14D7.

\bibliography{ref}
    \end{document}